\documentclass{article}
\usepackage{style}

\usepackage{times}

\usepackage{soul}
\usepackage{url}
\usepackage[hidelinks]{hyperref}
\usepackage[utf8]{inputenc}
\usepackage[small]{caption}
\usepackage{graphicx}
\usepackage{amsmath}
\usepackage{booktabs}
\usepackage{threeparttable}
\usepackage{amssymb} % Math notation
\usepackage{microtype} 

\usepackage{etoolbox}
\makeatletter
\patchcmd\@combinedblfloats{\box\@outputbox}{\unvbox\@outputbox}{}{%
   \errmessage{\noexpand\@combinedblfloats could not be patched}%
}%
 \makeatother

\title{Seq-U-Net: A One-Dimensional Causal U-Net for Efficient Sequence Modelling}

\author{
Daniel Stoller$^1$\footnote{Contact Author}\and
Mi Tian$^2$\and
Sebastian Ewert$^2$\and
Simon Dixon$^1$\\
\affiliations
$^1$Queen Mary University of London\\
$^2$Spotify London\\
\emails
d.stoller@qmul.ac.uk,
\{mtian,sewert\}@spotify.com,
s.e.dixon@qmul.ac.uk
}

\begin{document}

\maketitle

\begin{abstract}
Convolutional neural networks (CNNs) with dilated filters such as the Wavenet or the Temporal Convolutional Network (TCN) have shown good results in a variety of sequence modelling tasks.
However, efficiently modelling long-term dependencies in these sequences is still challenging.
Although the receptive field of these models grows exponentially with the number of layers, computing the convolutions over very long sequences of features in each layer is time and memory-intensive, prohibiting the use of longer receptive fields in practice.
To increase efficiency, we make use of the ``slow feature'' hypothesis stating that many features of interest are slowly varying over time.
For this, we use a U-Net architecture that computes features at multiple time-scales and adapt it to our auto-regressive scenario by making convolutions causal.
We apply our model (``Seq-U-Net'') to a variety of tasks including language and audio generation.
In comparison to TCN and Wavenet, our network consistently saves memory and computation time, with speed-ups for training and inference of over 4x in the audio generation experiment in particular, while achieving a comparable performance in all tasks.
\end{abstract}

\section{Introduction}

Sequence modelling is an important problem central to many application domains, including language, audio, and video generation~\cite{bai2018convolutional,yuSeqGANSequence2017,trinhLearningLongerterm2018}.
In some of these applications, the sequences can be hundreds of thousands of time-steps in length (e.g. in the case of audio generation due to the high sampling rate of audio signals), and it can be vital to model the long-term dependencies present in such sequences (for example to be able to repeat a melody in a music piece that occurred a minute earlier).

This problem is often framed as the task of predicting the next element in a sequence given all of the elements observed so far, giving rise to auto-regressive models.
In deep learning, recurrent neural networks (RNNs) are most commonly used as auto-regressive models, since they can theoretically remember inputs for an arbitrary number of time-steps, and also offer quick inference at test time as the hidden state carries all the information about previous sequence elements and only needs to be updated using the next element.
However, in practice, these models can be difficult~\cite{bengioLearningLongterm1994} and slow~\cite{trinhLearningLongerterm2018} to train due to their strictly sequential nature.
More recently, CNNs with dilated filters were shown to be competitive approaches for sequence modelling.
Instead of relying on recurrence to carry the relevant information through a hidden state over a large number of steps, which might be difficult to achieve in practice, these CNNs access far-away time-steps more directly through their dilated filters.
Notable examples include the temporal convolutional network (TCN)~\cite{bai2018convolutional} and Wavenet~\cite{dielemanWaveNetGenerative2016}.

Despite their impressive performance in a variety of tasks, these architectures suffer from two issues.
Firstly, each convolutional layer operates at the same time resolution as the input.
This results in a high memory usage and training time especially with long sequences, rendering long-term modelling infeasible even with large scale, multi-GPU training~\cite{dielemanWaveNetGenerative2016}.
Secondly, inference is slow as elements have to be predicted sequentially and require a forward pass through the CNN's many layers.
Re-using layer outputs from previous steps can mitigate some of these issues, yet even in this case all layers have to be traversed and updated to predict the next sequence element.

In this context, the ``slow feature analysis``~\cite{wiskottSlowFeature2002} hypothesis states that for a wide variety of tasks important features of an input signal vary only slowly over time -- which leads to an interesting approach of increasing efficiency by computing some features at lower sampling rates compared to the input without compromising model performance.
Notably, U-Nets~\cite{ronnebergerUnetConvolutional2015} already incorporate the equivalent of this principle for image processing, by computing features at different time-scales with two-dimensional convolutions and combining them to make predictions at the same resolution as the input.
A version with \emph{one-dimensional} convolutions was presented for audio source separation~\cite{stollerWaveUNetMultiScale2018}.
We base our model on this U-Net variant, as it should be able to process many kinds of temporal sequences, not just audio signals.
We show how to adapt it for our auto-regressive setting by making all convolutions causal, such that each prediction for the next time-step can only depend on past inputs.

As a result, we obtain the ``Seq-U-Net''\footnote{Code available at \mbox{\url{https://github.com/f90/Seq-U-Net}}}, a general-purpose network architecture that is not limited to audio tasks but can be applied to a wide range of sequence modelling problems -- while providing considerable efficiency improvements over TCN and Wavenet.
Inference is greatly accelerated by only computing new layer activations if they are not decimated in the downsampling process.
This time-variant processing gives each layer its own ``update rate'', which is in contrast to fully-convolutional TCN and Wavenet approaches.
In particular, we compare to TCN in the context of word- and character-level language modelling as well as symbolic music generation. 
Additionally, we tackle the task of generating piano music directly in the time-domain and compare performance with a Wavenet reimplementation using a log-likelihood metric as well as listening tests.
Overall, we find that our architecture achieves competitive results while requiring less memory and training time.

\begin{figure}[t]
  \centering
  \centerline{\includegraphics[width=\columnwidth]{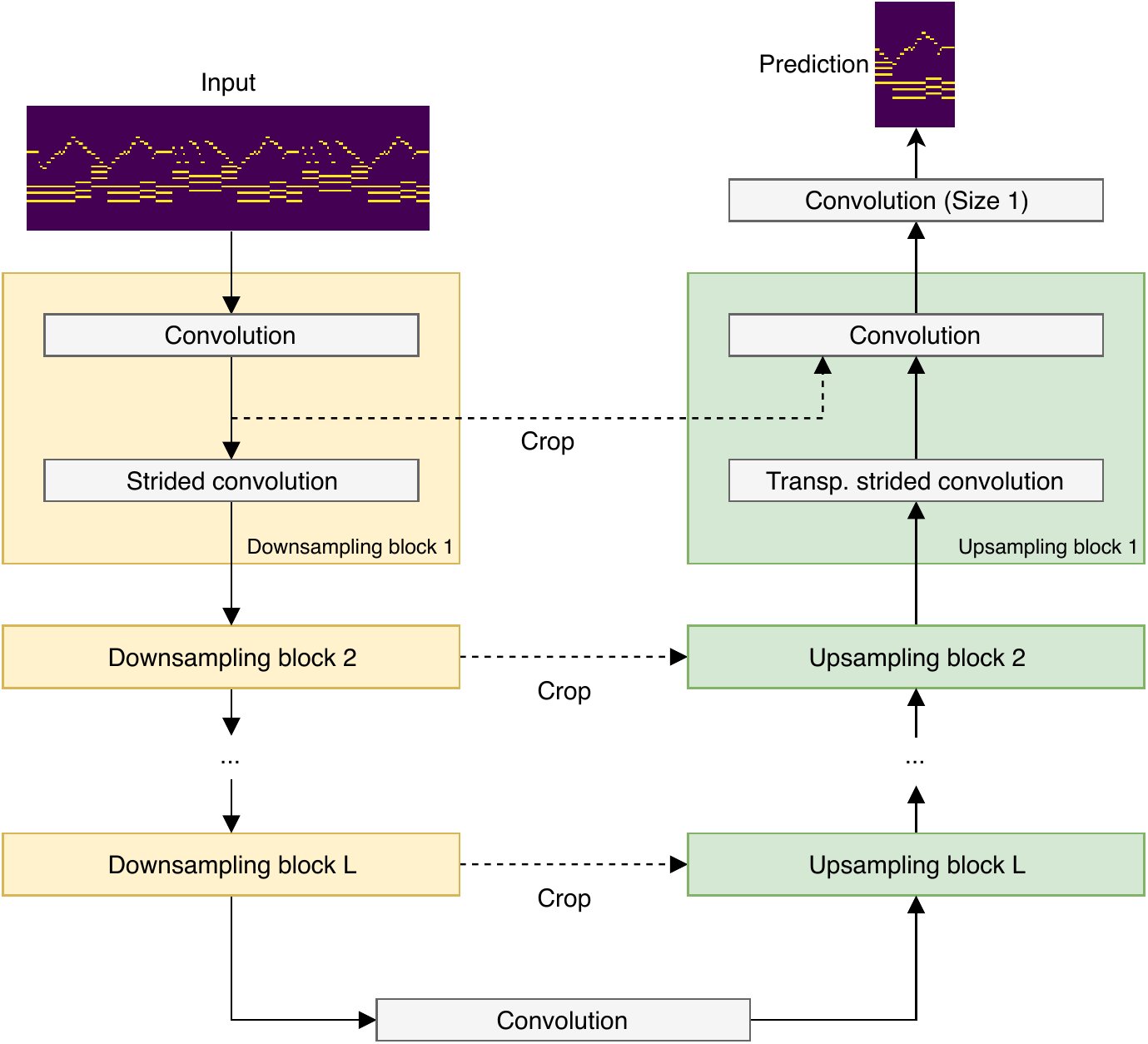}}
  \caption{Architecture diagram of our proposed model. Convolutions are one-dimensionally applied across time and include LeakyReLU activations followed by Dropout. Strided and transposed strided convolutions are used for down- and upsampling the features, respectively. Since the convolutions do not use padding, the output is smaller than the input and skip connections need to be cropped at the front.}
  \label{fig:architecture}
\end{figure}

\section{Related Work}

Recurrent neural networks (RNNs) are a commonly used approach in deep learning for sequence modelling, including LSTMs and GRUs~\cite{gravesSpeechRecognition2013,boulanger-lewandowskiModelingTemporal2012}.
In practice, training these models to successfully model long-term dependencies can be difficult~\cite{bengioLearningLongterm1994} and slow as computation is strictly sequential and can not be parallelised~\cite{trinhLearningLongerterm2018}.
Hierarchical multi-scale RNNs~\cite{chungHierarchicalMultiscale2016,elhihiHierarchicalRecurrent1995} and the Clockwork RNN~\cite{koutnikClockworkRNN2014} model time series on multiple time-scales to enable longer-term dependency modelling, but the sequential processing in the high-resolution timescales is still computationally expensive.
The SampleRNN~\cite{mehriSampleRNNUnconditional2016} is a three-layer RNN specifically developed for audio generation.
While it also employs a multi-scale approach to an extent, it inherits the disadvantages of RNNs mentioned above, and the ``slower'' layers have to compute high-level features directly from the raw audio input and forward them to the ``faster'' layers, which is arguably more difficult than computing them bottom-up using features from the ``faster'' layers.

Alternative approaches involve CNNs with filters that have increasing dilation factors to cover longer distances between inputs~\cite{kalchbrennerNeuralMachine2016,camposSkipRNN2017}, of which we want to highlight TCN~\cite{bai2018convolutional} and Wavenet~\cite{dielemanWaveNetGenerative2016} for sequence modelling.
Due to their depth, these neural models require a large amount of memory and have slow inference as a forward pass is required at each time-step.

The parallel Wavenet~\cite{oordParallelWaveNet2017} provides fast inference by using a flow-based student network to emulate the outputs of an already trained Wavenet.
For long-term dependency modelling in audio, \cite{dielemanChallengeRealistic2018} use a complex, multi-stage training with auto-encoding networks to compress the audio before using Wavenets to model the latent state evolution.
However, since these approaches involve training a Wavenet, they inherit its computational complexity.

Other approaches were developed such as FFTNet~\cite{jinFFTNetRealTime2018}, WaveRNN~\cite{kalchbrennerEfficientNeural2018} and MelNet~\cite{vasquezMelNetGenerative2019}, which do provide large efficiency gains by means of optimisations specific to the audio domain, but at the cost of generality.

Finally, the Transformer network~\cite{vaswaniAttentionAll2017} has shown great potential for sequence-based tasks, but the complexity of its attention mechanism is quadratic in the length of the sequence, preventing the use for long sequences.
Sparse Transformers~\cite{childGeneratingLong2019} restrict the attention modules to a sparse subset of all previous inputs to remedy this, but could still benefit from introducing a multi-scale architecture.

We are unaware of another multi-scale approach evaluated across a variety of sequence modelling problems, but similar approaches were used for video segmentation~\cite{shelhamerClockworkConvnets2016} and audio separation~\cite{stollerWaveUNetMultiScale2018}.

\section{Method}
\label{sec:method}

\begin{figure*}[t]
  \centering
  \centerline{\includegraphics[width=\textwidth]{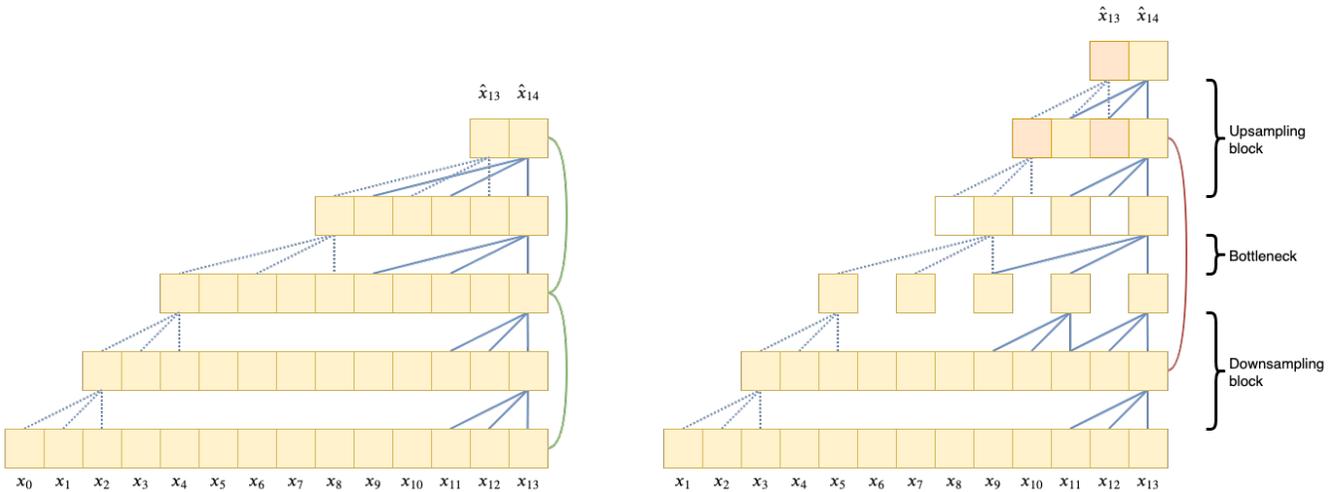}}
  \caption{Comparison between TCN (left, two residual blocks) and our proposed model (right, one down- and upsampling block). Due to the downsampling, features are computed only at certain regular intervals, saving memory and training time. Zero-padding is used in the upsampling blocks (white squares) to increase resolution again, leading to different computational paths throughout the network (red squares) High-resolution features that were computed earlier are combined in the upsampling block via cropping and concatenation (red line).}
  \label{fig:system_overview}
\end{figure*}

We present two variants of our multi-scale approach.
The first is an adaptation of the Wave-U-Net to the auto-regressive setting and shown in Section~\ref{sec:model_waveunet}.
The second variant, presented in Section~\ref{sec:model_residual}, further adds residual connections to stabilise training for tasks with very long-term dependencies such as raw audio generation.

\subsection{Seq-U-Net: A Causal Wave-U-Net}
\label{sec:model_waveunet}

Our model is based on the Wave-U-Net~\cite{stollerWaveUNetMultiScale2018} and shown in Figure~\ref{fig:architecture}.
The network features $L$ \emph{levels} of downsampling (DS) and upsampling (US) blocks, and a convolutional bottleneck and output layer.
Each downsampling block features a convolution, whose outputs are used as a shortcut connection for the respective upsampling block, followed by another convolution with stride $k$ to downsample the features across time.
Each upsampling block has a transposed convolution with stride $k$ to upsample the previously obtained coarse-grained features.
The result is concatenated with the features from the shortcut connection, and input to another convolution to combine high- and low-level features.
In this paper, we set the stride $k$ to $2$.
All convolutions have the same filter width and a LeakyReLU activation followed by Dropout, except for the output convolution.

Like in the original Wave-U-Net, the convolutional layers do not use zero-padding so that all model predictions are made with the necessary input context.
As a result, there are more feature frames in the shortcut output of a DS block than in the output of the transposed strided convolution in the corresponding US block.
In the Wave-U-Net, the outputs at each level of the network are interpreted as  features describing the center part of the input, so the shortcut features are center-cropped before concatenation.
Consequently, the source signals are predicted for the center part of the input mixture excerpt.

Our key idea is to interpret the filters as causal instead: the output of a filter covering input timesteps $n-k$ to $n+k$ should now help predict input $x_{n+k+1}$ \emph{outside} of its receptive field instead of some feature of the input at timestep $n$, i.e. the current source audio signal.
Therefore, we instead crop the first feature frames of each shortcut connection to make sure that features are aligned in time properly.
As a result, we obtain an auto-regressive model for sequence modelling, similar to Wavenet and TCN, but significantly sparser in terms of activations due to the decreased resolution in most of the layers.

\subsubsection{Fast inference}
\label{sec:method_inference}

From a signal processing perspective, TCN and Wavenet are time-invariant systems and apply the same set of operations at each time step. Time-invariant processing, however, is not required in autoregressive models. Therefore, inspired by \cite{koutnikClockworkRNN2014}, we combine our multiscale architecture with a time-variant processing scheme, which drastically accelerates our approach as many operations do not have to be computed at every step:
If an output computed for the latest time-step in a DS block is decimated (e.g. for the output $\hat{x}_{13}$ in Figure~\ref{fig:architecture}), only the US blocks on the same or higher resolution need to be updated, since the input to the other blocks does not change.
This means that a block on level $i \in \{1, \ldots, L\}$ only needs to be updated every $k^{i-1}$ time-steps.
To implement this inference procedure, all blocks are given an internal clock based on their level to determine when to compute a new output.
To predict the very first sample from a given context, a normal forward-pass is conducted, including initialising the caches for the activations in each layer, before switching to the above step-wise procedure. 
For further details on the implementation, please refer to our source code.

\subsection{Residual variant}
\label{sec:model_residual}

Since audio generation involves especially long-term dependencies, we employ much deeper instances of our model for the experiment in Section~\ref{sec:experiments_audio} to increase its receptive field.
With this increase in layers however, we observed instability during the network training.
Residual networks can be trained stably even in the presence of hundreds of layers~\cite{heDeepResidual2015}, so we also propose a residual variant of our model.

Compared to the baseline model from Section~\ref{sec:model_waveunet}, we employ an additional convolution on the input with $F$ output channels, and also use $F$ input and output channels for all up- and downsampling blocks to allow for residual connections.
We replace each convolutional layer in the base model with a residual layer similar to the one in Wavenet~\cite{dielemanWaveNetGenerative2016}, whose outputs $y$ are given by

\begin{equation}
    y = I(x) + \mathrm{tanh}(C_1(x)) \cdot \sigma(C_2(x)),
\end{equation}

where $x$ are the layer inputs, $\sigma$ is the sigmoid function, $C_i$ applies convolutional layer $i$ to its input (which includes biases) and $I$ processes the input $x$ to provide an identity connection in case the convolutions change the feature dimensionality.
Dropout was omitted in this variant since overfitting was not a large concern in our audio generation experiments, but could readily be added to the residual convolutions.

For the convolutions with stride used in the DS blocks, $I$ involves decimating the input $x$ to provide the identity for the residual layer. 
For the transposed convolutions with stride in the US blocks, $I$ takes the input and repeats the feature vector at each time step $k-1$ times to perform upsampling\footnote{Note that this is also a causal operation that does not violate the auto-regressive condition.}.
For both down- and upsampling, $I$ finally crops the resulting feature sequence at the front to ensure it matches the number of residual features, which is reduced due to not using padding for convolutions.
To refine the high-resolution shortcut features using the low-resolution features from the upsampling path, we use the shortcut as input $x$ and use the concatenation of the shortcut and the upsampled features as input to the residual convolutions $C_i$.

To easily scale the network in size for more complex tasks, we also add a depth parameter $D$ that results in $D+1$ residual layers in each down- and upsampling block (one additional layer in each block for up- or downsampling), allowing features to be processed more flexibly at each time resolution. 

\section{Complexity analysis}
\label{sec:complexity}

We will analyse the memory consumption and computational complexity of our approach at both training and test time and compare with Wavenet\footnote{Comparison with TCN is omitted as it is very similar to Wavenet but differs slightly in the number of layers per level of resolution}.

\subsection{Training}
\label{sec:complexity_training}

Due to the size $N$ of receptive field increasing exponentially with the number of layers for the Seq-U-Net and Wavenet, roughly $L = \log_k(N)$ levels of processing are required. 
For the Wavenet, we define $k$ as the factor with which dilation increases in each layer.

When presented with $I \geq N$ inputs during training, Wavenet needs to compute $I$ feature activations in each of the $L$ layers, since it operates on the same resolution as the input, reaching a total of $I \cdot \log_k(N)$. 
The Seq-U-Net on the other hand computes $3I+\frac{I}{k}$ feature activations in the first down- and upsampling block, $3\frac{I}{k}+\frac{I}{k^2}$ on the second level, and so on, in addition to a bottleneck convolution with $\frac{I}{k^L}$ outputs.
For the Seq-U-Net, we thus obtain at most $\sum_{i=0}^{L} 4\frac{I}{k^i} \leq 8I$ feature activations regardless of the number of layers\footnote{This disregards the reduction in size due to not using padding for convolutions since it occurs in all models}.
The above calculation not only demonstrates the time complexity, but also the required memory, since the computed feature activations need to be maintained for the backward-pass.

\subsection{Inference}
\label{sec:complexity_inference}

To generate a sequence at test time with auto-regressive models such as Wavenet and Seq-U-Net, elements have to be predicted one by one, conditioned on the previously generated ones. 
As a result, the time required for one sample is important, especially when sequences are long (e.g. in audio generation) or when a real-time application is desired.
While caching previously computed outputs in the Wavenet reduces computation time, it still involves evaluating all $L$ residual layers.
This presents a significant burden as Wavenet implementations tend to be deep (also noted in~\cite{kalchbrennerEfficientNeural2018}, where the authors used $L=30$).
Thus, we require at least $L$ convolution operations per time-step.

In the Seq-U-Net however, each level in the network only has to be updated at certain intervals as described in Section~\ref{sec:model_waveunet}.
In particular, the average number of levels we have to update for each time-step is $\sum_{i=1}^L \frac{1}{k^{i-1}} \leq 2$ and thus a constant number of layers independent of the number of levels $L$ in the network.
While this is an amortised analysis of the average time per step, in the worst case all layers need to be updated, although this is not of practical importance when generating longer sequences offline.

\section{Experiments}

\begin{table}[t]
\footnotesize
\centering
\begin{threeparttable}
\setlength{\tabcolsep}{1.15ex}
\begin{tabular}{cccccc}
\toprule
Task & Model & Train & Test & Time (s) & Mem. \\
\midrule
Char-LM & TCN & 1.066 & 1.31 & 0.0694 & 445.9 \\
Char-LM & Seq-U-Net & 1.08 & 1.30 & 0.0286 & 304.9 \\
\midrule
Word-LM & TCN & 47.21 & 108.47 & 0.0480 & 580.5 \\
Word-LM & Seq-U-Net & 40.43 & 107.95 & 0.0234 & 382.1 \\
\midrule
M-Muse & TCN & 5.789 & 6.931 & 0.0059 & 108.5 \\
M-Muse & Seq-U-Net & 5.794 & 6.969 & 0.0065 & 75.3 \\
\midrule
M-Nott & TCN & 1.409 & 2.783 & 0.0071 & 73.1 \\
M-Nott & Seq-U-Net & 1.850 & 2.97 & 0.0067 & 52.5 \\
\midrule
M-JSB & TCN & 6.178 & 8.154 & 0.0034 & 13.1 \\
M-JSB & Seq-U-Net & 6.151 & 8.173 & 0.0037 & 8.2 \\
\midrule
Piano & Wavenet & 1.76 & 1.88 & 1.4616* & 5294* \\
Piano & Seq-U-Net** & 1.83 & 1.93 & 0.3621* & 1514* \\
\bottomrule
\end{tabular}
\begin{tablenotes}
    \footnotesize
    \item[*]{Measurements were taken with a batch size of $2$ instead of $16$ due to the high amount of memory required.}
    \item[**]{Residual variant}
\end{tablenotes}
\end{threeparttable}
\caption{Performance metrics for our model and TCN and Seq-U-Net comparison models across the different tasks. \mbox{``M-''} indicates a symbolic music modelling task. Times denote the duration for a forward- and backward-pass, averaged over a whole epoch. ``Mem.'' indicates maximum memory consumption of tensors within GPU memory during one training epoch in MB, excluding cache and other sources of memory usage.}
\label{tab:results}
%\vspace{-0.2cm}
\end{table}

We evaluate our method on a variety of sequence modelling tasks regarding its performance, training time and memory complexity.
Due to the architectural similarity, we will firstly compare our method with TCN in Section~\ref{sec:comparison_tcn} on language modelling as well as symbolic music modelling.
To test whether our model can capture very long-term dependencies, we finally perform audio generation in the time-domain and compare to a Wavenet baseline.

For time and memory measurements, we use a single NVIDIA GTX 1080 GPU with Pytorch 1.2, CUDA 9 and cuDNN 7.5\footnote{We use Pytorch's benchmark mode to find the best algorithm for training each network.}.
We compare the average time required for each training step\footnote{Includes the forward-pass through the network for one batch, the loss and gradient calculation and the the weight update, but not the overhead of batch preparation}, and the maximum memory allocated throughout a training epoch\footnote{Does not include memory used for purposes such as caching}.

\subsection{Comparison with TCN}
\label{sec:comparison_tcn}

We will compare our model against TCN across three sequence modelling tasks.
To match model complexity, we use the same filter length, Dropout rate, and levels of resolution, which results in very similar receptive field size.
Then, the number of features in each layer is adapted for Seq-U-Net so it matches TCN in the number of parameters.

We optimise each model for $100$ epochs using a batch size of $16$ and an Adam optimiser with initial learning rate $\alpha$, which is reduced by half if validation performance did not improve after $P$ epochs and more than $10$ epochs have passed since the beginning of training.
Finally, the model that performed best on the validation set is selected.

To prevent the training procedure from favouring one model over the other, we perform a hyper-parameter optimisation over the learning rate $\alpha$ in the range of $[e^{-12}, e^{-2}]$ and optional gradient clipping with magnitudes between $[0.01,1.0]$.
This hyper-parameter optimisation is performed for each combination of model and task using a tree of Parzen estimators\footnote{``Hyperopt'' package: \mbox{\url{http://hyperopt.github.io/hyperopt/}}} to find the minimum validation loss.
All hyper-parameters are shown in Table~\ref{tab:hyperparams}.

\subsubsection{Character-based language modelling}
\label{sec:experiments_char}

We perform character-based language modelling, where the task is to predict the next character given a history of previously observed ones, on the PTB dataset~\cite{marcusBuildingLarge1993}.
The average cross-entropy loss is used as training objective, and patience is set to $P=5$.

For both models, we use $100$-dimensional character embeddings with $0.1$ Dropout as input, and their output is projected back to character probabilities using the transposed version of the embedding matrix.
We evaluate models using the bits-per-character (bpc) metric.

As shown in Table~\ref{tab:results}, our model performs as well as its TCN counterpart in this regard, while requiring 59\% less time per training step, and 32\% less GPU memory during training.
These results suggest that many of the required features are on a higher level of abstraction and vary only slowly, e.g. per word or per sentence, and so do not need to be recomputed for each new character -- a hypothesis also put forth in~\cite{chungHierarchicalMultiscale2016}.

\subsubsection{Word-based language modelling}
\label{sec:experiments_word}

For our second experiment, we perform word-based language modelling, which involves predicting the next word following a given sequence of words.
As in the previous experiment, we use the PTB dataset with a vocabulary of 10,000 words.
Following TCN's experimental set-up~\cite{bai2018convolutional}, we use $600$-dimensional word embeddings with $0.25$ Dropout as input, and use the transpose of the embedding matrix to project the $600$-dimensional outputs from the models to probability vectors over all words. 
For training, we minimise the average cross-entropy with a patience of $P=5$, and for evaluation we use the per-word perplexity.

Similarly to the results for character-based language modelling in Section~\ref{sec:experiments_char}, Table~\ref{tab:results} shows that both models perform very similarly, but the Seq-U-Net architecture is substantially more efficient to train (reducing the training time by 51\% and memory usage by 34\%).

\subsubsection{Symbolic music modelling}
\label{sec:experiments_music}

For our final comparison with TCN, we model polyphonic music in the symbolic domain.
Each music piece is represented as a piano roll -- a binary matrix of size $88 \times T$ that indicates which of the $88$ pitches are active at each of the $T$ time frames.
For simplicity, we assume that pitch activations at a given time frame are independent of each other\footnote{This assumption is commonly used in music transcription, see e.g.~\cite{ycartPolyphonicMusic2018}}.
This allows our models to predict a whole time-frame at each step in an auto-regressive manner, and using the sum of binary cross-entropies over each pitch, averaged over all time frames as training objective.
We use a patience of $P=10$ for early stopping.

Three different datasets of varying complexity and content are used: Muse\footnote{See \url{http://www-etud.iro.umontreal.ca/~boulanni/icml2012}}, Nottingham (Nott)\footnote{See \url{http://ifdo.ca/~seymour/nottingham/nottingham.html}} and the JSB chorales~\cite{allanHarmonisingChorales2005}.
For evaluation, we use the frame-wise perplexity introduced in~\cite{boulanger-lewandowskiModelingTemporal2012}.

Table~\ref{tab:results} shows the perplexity on the training and test sets for both models on all datasets.
We find that both models are very closely matched in terms of training and test perplexity on the Muse and JSB datasets.
For the Nott dataset, TCN achieves a noticeably lower perplexity than the Seq-U-Net on the training partition.
This performance gap also appears on the test set, although it is considerably smaller, indicating that incorporating the slow feature hypothesis induces a regularising effect on the model as it encourages some features to be slowly varying.

For all datasets, the improvement in training time is not as pronounced compared to the previous language modelling experiments.
This is due to the much smaller size of the models, where the higher number of convolutional layers in the Seq-U-Net has a larger impact than the reduction in computation time for each layer.
Nevertheless, the memory footprint is substantially reduced by an average 32\%.

\subsection{Raw audio generation}
\label{sec:experiments_audio}

\begin{table}[t]
%\footnotesize
\centering
\begin{tabular}{cccccc}
\toprule
Model & Layers & Features & Context & Width \\
\midrule
Wavenet & 13 & 128 & 32764 & 2 \\
Seq-U-Net & 11 & 180 & 32748 & 5 \\
\bottomrule
\end{tabular}
\caption{Models used for audio generation. Context is given as a number of audio samples, and width describes the convolutional filter width in each layer.}
\label{tab:audio}
%\vspace{-0.2cm}
\end{table}

To test whether our model can capture long-term dependencies in very long, complex sequences, we apply it to the generation of audio waveforms, using the residual variant presented in Section~\ref{sec:model_residual}.
Since our architecture resembles the Wavenet with its use of stacks of residual convolutions, we use it as our comparison model in the following.

In particular, we use the classical piano recordings as used in~\cite{dielemanChallengeRealistic2018} amounting to about 607 hours in duration, and partition them into a training and test set, while avoiding pieces overlapping between the two partitions.
Note that our version of the dataset is different as we were not able to obtain all the recordings listed in~\cite{dielemanChallengeRealistic2018}.

We train two models in this experiment, listed in Table~\ref{tab:audio}.
The first one is a Wavenet baseline comprised of 4 Wavenet stacks with 13 dilated convolutional layers each
and 512 features in the skip connection, and the second one is a Seq-U-Net model that matches the Wavenet in terms of receptive field size, and uses a residual depth of $D=2$.

Besides downsampling the audio to 16 KHz mono signals, no further preprocessing is applied.
During training, audio excerpts are loaded from random positions within the audio files, and each audio sample is transformed into a $256$-dimensional one-hot vector using 8-bit mu-law encoding, following the Wavenet approach~\cite{dielemanWaveNetGenerative2016}.
A training batch consists of 16 examples and uses the last 5000 audio samples in each example as simultaneous training targets for the model. 
The average cross-entropy is minimised over $246000$ iterations (equivalent to just over one epoch) with an Adam optimiser and a learning rate of~$0.0005$. 

\subsubsection{Experimental setup}

For evaluation, we report the likelihood of the models in bits per audio samples (bpa) on the test set. 
However, the bpa metric might not reflect perceptual audio quality very well, especially since the model uses its own predictions as input and not real samples at test time.
This discrepancy is well known in the literature~\cite{huszarHowNot2015}, and we also found in practice that the two models vary in their stability at generation time.
While the Wavenet starts to introduce progressively more noise into its outputs with longer generation, the Seq-U-Net appears stable throughout.
Since this effect is very pronounced with durations of 10 seconds or longer, making Seq-U-Net clearly preferable, we conducted a listening test with samples of 5 seconds.
We used a temperature of $0.95$, meaning the unnormalised model outputs were divided by $0.95$ before applying the softmax to obtain probabilities, making model predictions more ``conservative''.
In preliminary experiments, we found this stabilises the generation process, resulting in increased quality for both models.

Each of the $20$ questions presented the participant with a 1.5 second excerpt of real piano randomly sampled from our test dataset.
This was followed by two continuations produced by our two models that also include the real excerpt in the beginning.
This conditional generation setting allows directly comparing between outputs of different models for the same input context:
The participants were asked which excerpt has ``better timbre (does it sound like a piano, is the audio free of distortions?)''  and ``more musical coherence (with respect to melody, harmony, rhythm)''. 
An additional ``Not sure'' option was available when the participant is unsure or thinks the quality is the same for both excerpts.
The total number of participants is $22$.

\subsubsection{Results}

\begin{figure}[t]
  \centering
  \centerline{\includegraphics[width=\columnwidth]{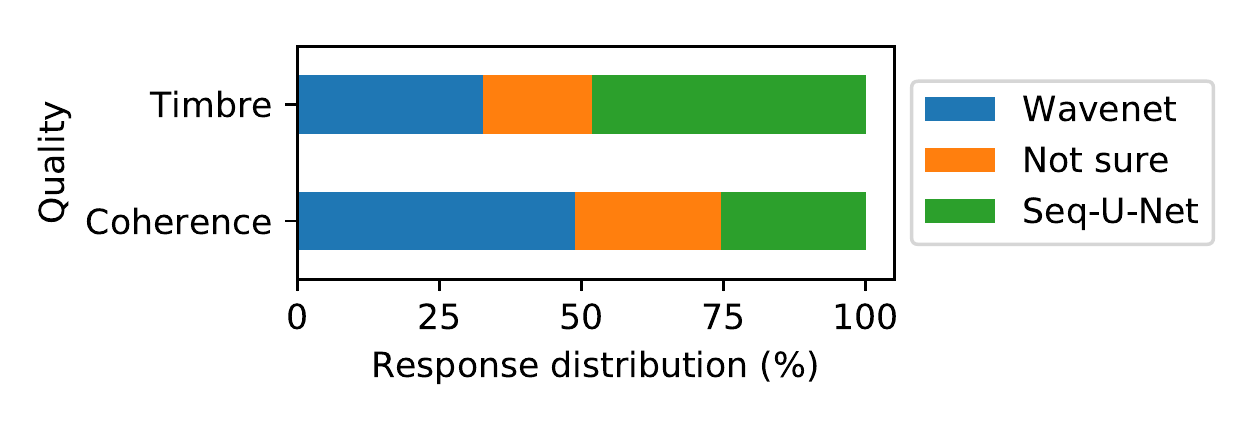}}
  \caption{Results of the listening test, showing the overall distribution of responses for both the timbre and the musical coherence questions}
  \label{fig:listening_test}
\end{figure}

As seen in Table~\ref{tab:results}, the Wavenet slightly outperforms the Seq-U-Net in terms of the bpa metric, albeit achieving a small relative improvement of $2.6\%$ on the test set, indicating the models are closely matched in terms of performance.
The training set results indicate this might be due to the Wavenet fitting the training set more closely in the given number of training iterations.
At the same time, the required training time and memory are drastically reduced for the Seq-U-Net by a factor of $4$ and $3.5$, respectively.

\begin{table*}[t]
%\footnotesize
\centering
\begin{tabular}{ccccccccccc}
\toprule
Task & Model & $W$ & $L$ & $H$ & Dropout & Context & Params & $P$ & LR & Clip \\
\midrule
Char-LM & TCN & 3 & 4 & 600 & 0.1 & 80 & 5.9M & 5 & 0.00014 & 0.213 \\
Char-LM & Seq-U-Net & 3 & 4 & 390 & 0.1 & 73 & 5.9M & 5 & 0.00073 & No \\
\midrule
Word-LM & TCN & 3 & 4 & 600 & 0.5 & 73 & 14.7M & 5 & 0.00115 & No \\
Word-LM & Seq-U-Net & 3 & 4 & 390 & 0.5 & 73 & 14.9M & 5 & 0.00037 & 0.722 \\
\midrule
Music-Muse & TCN & 5 & 4 & 215 & 0.2 & Full & 1.7M & 10 & 0.00023 & No \\
Music-Muse & Seq-U-Net & 5 & 4 & 150 & 0.2 & Full & 1.7M & 10 & 0.00047 & No \\
\midrule
Music-Nott & TCN & 5 & 4 & 215 & 0.2 & Full & 1.7M & 10 & 0.000067 & 0.601 \\
Music-Nott & Seq-U-Net & 5 & 4 & 150 & 0.2 & Full & 1.7M & 10 & 0.00108 & No \\
\midrule
Music-JSB & TCN & 3 & 2 & 220 & 0.5 & Full & 534k & 10 & 0.00134 & No \\
Music-JSB & Seq-U-Net & 3 & 2 & 170 & 0.5 & Full & 522k & 10 & 0.00051 & 0.324 \\
\bottomrule
\end{tabular}
\caption{Hyper-parameters used for TCN and Seq-U-Net comparisons. $H$ is the number of filters in each convolutional layer, LR and Clip are the best learning rate and clipping magnitude found through hyper-parameter optimisation, and $W$ the convolutional filter width.}
\label{tab:hyperparams}
\end{table*}

The results of the listening test are shown in Figure~\ref{fig:listening_test}.
While the Seq-U-Net exhibits better timbral characteristics, producing better continuations than the Wavenet in 15 out of the 20 provided examples, it falls behind in terms of musical coherence.
We suspect this is due to the Seq-U-Net sometimes producing an unexpected transition from the real excerpt to the generated section, but then producing sounds more stably as time goes on.
Overall, the two models appear to have different strengths and weaknesses -- we encourage the reader to listen to the audio examples provided in the supplementary material.
Additionally, the high amount of "Not sure" responses, especially for such a sensitive paired discrimination task, indicates that the models are quite evenly matched in this setting.

Finally, we measure the performance impact of our inference method introduced in Section~\ref{sec:method_inference} by comparing to the Wavenet's generation speed when caching previous activations.
With a batch size of $1$ on a single NVIDIA GTX 1080 GPU, we achieve $69$ audio samples per second for the Wavenet, and $309$ for the Seq-U-Net and thereby a speed-up with a factor greater than $4$.

\section{Discussion}

The predictive performance of the Seq-U-Net as outlined in Table~\ref{tab:results} is remarkably similar to that of Wavenet and TCN comparison models across all tasks we tested.
While efficiency gains are not very noticeable for very small instances of our model with few levels of resolution, they rapidly increase when moving towards larger and deeper models as used in language and audio modelling, and we can expect these gains to become more pronounced for even deeper models with even longer receptive fields.

Since the metrics used in Table~\ref{tab:results} are based on how much probability the models assign to the test data (log-likelihood) and not directly on how realistic their generated output is, we performed a listening test for the piano audio generation task.
Surprisingly, despite better log-likelihood, our implementation of the Wavenet accumulates noise during generation, making it unsuitable to generate longer music pieces, whereas the Seq-U-Net is stable but less capable of smoothly continuing the real excerpts, for reasons that remain unclear.
A more unified approach to training and evaluating generative models would be desirable, so models can be more directly adapted for stability during generation time, instead of relying on architecture choices alone to ensure stability.

\section{Conclusion}

In this paper, we demonstrated how a causal variant of a U-Net architecture with one-dimensional convolutions across the time domain can perform on par with existing state-of-the-art models in a variety of sequence modelling tasks, while significantly reducing training time and memory requirements.
This is achieved by relying on the hypothesis that many relevant features in real-world sequences are only slowly varying over time, allowing the use of convolutional layers that compute features at progressively lower resolutions.

These efficiency gains make it feasible to train generative models with much longer receptive fields in the future, which can be very useful in domains such as music and language generation and is left for future work.

One limitation of our approach is that the levels of resolution along with the processing capacity at each resolution has to be manually pre-defined, which could limit performance.
Future work could include potential solutions as used in the Phased LSTM~\cite{neilPhasedLSTM2016} so the model can adapt its levels of resolution more dynamically to the task at hand.

Finally, attention mechanisms have shown great potential for sequence modelling and could be integrated into our approach by using attention operations in each down- and upsampling block alongside or instead of convolutions to further improve performance, as suggested in~\cite{childGeneratingLong2019}.

\bibliographystyle{named}
\bibliography{Research}

\end{document}